\documentclass[11pt]{article}

\usepackage[final]{acl}

\usepackage{times}
\usepackage{latexsym}

\usepackage[T1]{fontenc}

\usepackage[utf8]{inputenc}

\usepackage{microtype}

\usepackage{inconsolata}

\usepackage{graphicx}
\usepackage{booktabs}
\usepackage{amsmath}
\usepackage{amssymb}
\usepackage{multirow}
\usepackage{xcolor}
\usepackage{colortbl}
\usepackage{subcaption}
\usepackage{algorithm}
\usepackage{algorithmic}
\usepackage{tikz}
\usepackage{pgfplots}
\usepackage{listings}
\usepackage[most]{tcolorbox} 
\usepackage{enumitem}
\usepackage{xspace}

\setlist[itemize]{nolistsep,noitemsep,leftmargin=*}
\pgfplotsset{compat=1.16}
\usetikzlibrary{shapes,arrows,positioning,fit,backgrounds}

\definecolor{gpt5color}{RGB}{74,144,226}
\definecolor{deepseekcolor}{RGB}{80,200,120}
\definecolor{minimaxcolor}{RGB}{255,165,0}
\definecolor{kimicolor}{RGB}{186,85,211}
\definecolor{geminicolor}{RGB}{255,99,71}
\definecolor{gptosscolor}{RGB}{128,128,128}

\newtcolorbox{paperbox}[2][]{%
  enhanced,
  colback=white,
  colframe=black,
  fonttitle=\bfseries,
  before=\refstepcounter{table},
  title={#2},
  #1
}

\lstdefinelanguage{json}{
  morestring=[b]",
  morecomment=[l]{//},
  showstringspaces=false
}

\newcommand{\method}{MTRouter\xspace}
\newcommand{\github}{\url{https://github.com/ZhangYiqun018/MTRouter}\xspace}

\title{MTRouter: Cost-Aware Multi-Turn LLM Routing  with History–Model Joint Embeddings}

\author{
  \textbf{Yiqun~Zhang}$^{1~2}$ \quad 
  \textbf{Hao~Li}$^2$  \quad
  \textbf{Zihan~Wang}$^1$ \quad
  \textbf{Shi~Feng}$^{1~\dag}$
  \textbf{Xiaocui~Yang}$^1$ \quad \\
  \textbf{Daling~Wang}$^1$ \quad 
  \textbf{Bo~Zhang}$^2$ \quad 
  \textbf{Lei~Bai}$^{2}$ \quad
  \textbf{Shuyue~Hu}$^{2~\dag}$\\
  $^1$ \text{School of Computer Science and Engineering, Northeastern University} \\
  \text{Shenyang 110819, China} \\
  $^2$ \text{Shanghai Artificial Intelligence Laboratory} \\
  \texttt{yiqunzhang@stumail.neu.edu.cn} \\
  \texttt{\{fengshi,yangxiaocui,wangdaling\}@cse.neu.edu.cn} \\ 
  \texttt{\{lihao4,zhangbo,bailei,hushuyue\}@pjlab.org.cn}
}

\begin{document}
\maketitle
	\begin{abstract}
Multi-turn, long-horizon tasks are increasingly common for large language models (LLMs), but solving them typically requires many sequential model invocations, accumulating substantial inference costs. Here, we study cost-aware multi-turn LLM routing: selecting which model to invoke at each turn from a model pool, given a fixed cost budget. We propose MTRouter, which encodes the interaction history and candidate models into joint history–model embeddings, and learns an outcome estimator from logged trajectories to predict turn-level model utility. Experiments show that MTRouter improves the performance–cost trade-off: on ScienceWorld, it surpasses GPT-5 while reducing total cost by 58.7\%; on Humanity’s Last Exam (HLE), it achieves competitive accuracy while reducing total cost by 43.4\% relative to GPT-5, and these gains even carry over to held-out tasks. Further analyses reveal several mechanisms underlying its effectiveness: relative to prior multi-turn routers, MTRouter makes fewer model switches, is more tolerant to transient errors, and exhibits emergent specialization across models.
	Code: \github.
	\end{abstract}

\def\thefootnote{\dag}\footnotetext{Corresponding author.}\def\thefootnote{\arabic{footnote}}

\section{Introduction}

Large Language Models (LLMs) are increasingly deployed to solve complex, tool-using tasks that require extended sequences of interactions, such as software engineering \citep{jimenez2024swebench,jain2024livecodebench} and multi-step reasoning \citep{wang2022scienceworld, yang2024sweagent,team2025tongyi,phan2025humanitysexam}. A defining characteristic of these LLMs is their long-horizon nature, often requiring dozens of sequential model calls per episode. As these trajectories grow, the cumulative inference cost—exacerbated by expanding context windows~\cite{dao2023flashattention} and superlinear token consumption \citep{gao2025lessempiricalstudyturncontrol}—becomes a primary barrier to practical deployment and reproducible research.

\begin{figure}[!tbp]
\centering
\includegraphics[width=\linewidth]{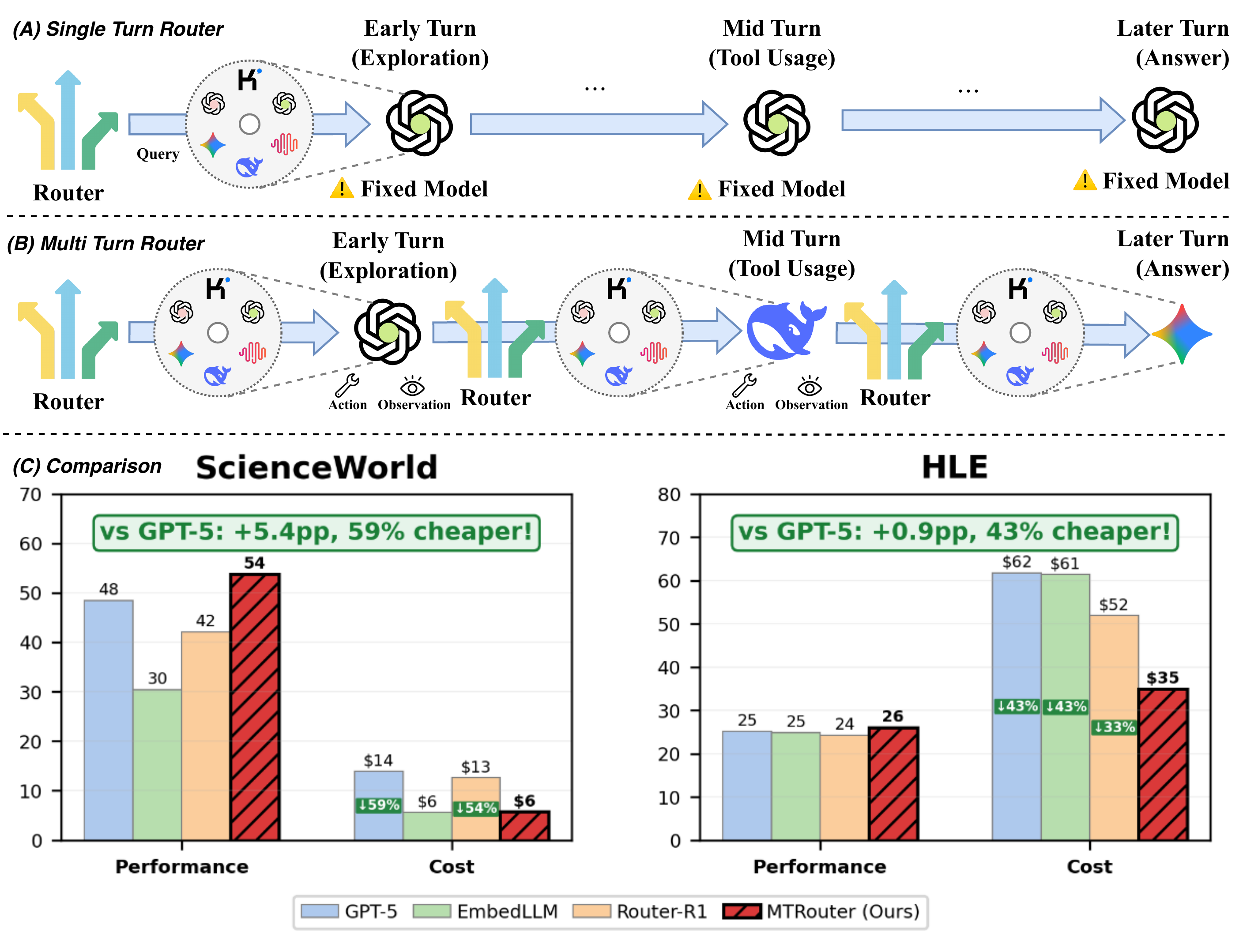}
\caption{\textbf{Top:} a single-turn (episode-level) router selects one model and keeps it fixed throughout the episode. \textbf{Middle:} a multi-turn router can adapt the model choice across turns based on the changing interaction state. \textbf{Bottom:} \method achieves a better performance--cost trade-off than representative baselines on ScienceWorld~\citep{wang2022scienceworld} and Humanity's Last Exam (HLE)~\citep{phan2025humanitysexam}.}
\label{fig:toy_example}
\vspace{-1.0em}
\end{figure}

This economic pressure is framed by a stark disparity in the current model landscape: frontier models provide state-of-the-art reasoning (like \texttt{claude-opus-4.5}\footnote{\url{https://www.anthropic.com/claude/opus}}, \texttt{gpt-5\footnote{\url{https://openai.com/index/gpt-5-system-card}}}) but are often two orders of magnitude more expensive than their lightweight counterparts (like \texttt{DeepSeek-v3.2~\citep{liu2025deepseek}, Kimi-K2~\citep{team2025kimi}}). To mitigate costs, prior work has explored single-turn (episode-level) routing, which selects a single model at the start of an episode based on the initial prompt (Figure~\ref{fig:toy_example}, Top). However, this static approach is inherently sub-optimal for long-horizon tasks. A single episode often consists of heterogeneous steps, ranging from high-stakes strategic planning to routine tool invocations and data formatting. Using a frontier model for every turn is wasteful, yet relying solely on a cheap model risks catastrophic failure during critical junctures.

This observation motivates multi-turn (turn-level) routing, where the agent adaptively switches between models at each step of the interaction (Figure~\ref{fig:toy_example}, Middle). While promising, multi-turn routing poses a significant predictive challenge: a router must assess whether a specific model choice at the current turn will jeopardize the final outcome dozens of steps later. Simple heuristics or localized error detection are often insufficient to capture the long-term impact of a model's performance on the entire trajectory.

We propose \textbf{\method}, which learns an \emph{outcome estimator} from logged trajectories.
The estimator maps a history--model pair to an estimate of the eventual episode outcome (terminal score/accuracy), using an error-aware adjustment to provide a stable training signal from offline data.
At inference time, the router selects the candidate model with the highest predicted outcome at each turn, and the episode is evaluated under fixed cost and turn limits.

Empirically, \method delivers consistent gains in both performance and cost.
On ScienceWorld (test), it improves average score from 48.4 (GPT-5) to 53.8 while reducing total cost by 58.7\%; on HLE (test), it reaches competitive accuracy while reducing total cost by 43.4\%.
It also generalizes under semantic distribution shift (OOD), maintaining improved outcomes with substantial cost savings (Table~\ref{tab:combined_results}).
Beyond aggregate metrics, our analysis shows that multi-turn routing is not ``switch more'': \method reaches success with fewer model switches than Router-R1~\cite{zhang2025router}, is more tolerant to transient errors, and exhibits structured model usage and emergent specialization across tools/actions (Figures~\ref{fig:cost_switches}--\ref{fig:action_by_model}).

Our key contributions are as follows:

\begin{itemize}[nolistsep, noitemsep, leftmargin=*]
\item We introduce \textbf{\method}, which learns an outcome estimator over history--model pairs from offline trajectories and performs turn-level routing in multi-turn agent episodes.
\item We evaluate on ScienceWorld and HLE (test and OOD) with a fixed candidate pool, and show consistent improvements in both performance and total cost over strong routing baselines, including Router-R1 and a representative commercial router.
\item We provide analyses that connect these gains to concrete routing behaviors, including fewer unnecessary switches, improved error recovery, and emergent specialization across tools/actions.
\end{itemize}

\section{Related Work}
\paragraph{LLM Routing.}
The proliferation of LLMs with diverse cost-capability profiles has motivated research on intelligent model selection. FrugalGPT \citep{chen2023frugalgpt} cascades models from cheap to expensive until confidence thresholds are met. EmbedLLM \citep{wang2024embedllm} learns embeddings to predict model performance on specific queries. 
RouterDC \citep{chen2024routerdc} uses dual contrastive learning for effective router training.
Avengers~\cite{zhang2025avengers} uses a simple clustering-based routing scheme to orchestrate a pool of ten 7B models, achieving performance that surpasses GPT-4.1, while AvengersPro~\cite{zhang2025beyond} extends this design to deliver Pareto-optimal performance--cost trade-offs under balanced evaluation settings.
These approaches focus on \emph{single-turn} routing: given a query, select one model to answer it. 
Moving towards multi-turn scenarios, Router-R1 \citep{zhang2025router} trained a policy LLM to interleave reasoning and routing through reinforcement learning. Unlike Router-R1 which relies on a heavy LLM-based router, our method learns a \textit{lightweight} outcome estimator over joint history-model embeddings, enabling cost-efficient turn-level routing.

\paragraph{LLMs Tool Use.}
LLMs augmented with tools can perform complex multi-step tasks \citep{schick2023toolformer,qin2023toolllm}. ReAct \citep{yao2023react} interleaves reasoning and action within a single prompt, while Toolformer \citep{schick2023toolformer} fine-tunes models to invoke tools autonomously. Recent studies have explored LLM-based tool use across diverse domains, including software engineering \citep{jimenez2024swebench,yang2024sweagent}, web browsing \citep{zhou2024webarena}, operating system interaction \citep{xie2024osworld}, complex reasoning \citep{mialon2023gaiabenchmarkgeneralai}, and autonomous research \citep{team2025tongyi}. These tasks are inherently interactive and typically cannot be resolved in a single turn, necessitating repeated action-observation cycles.
While most tool-use frameworks rely on a fixed underlying model, our work introduces a routing layer that dynamically switches between models, treating the model selection as an adaptive decision at each step. This approach is complementary to existing tool-use methodologies and can be generalized to multi-turn LLM system.

\section{MTRouter}
\label{sec:method}


\subsection{Problem Formalization}
\label{sec:formalization}

\begin{figure*}[!t]
\centering
\includegraphics[width=\linewidth]{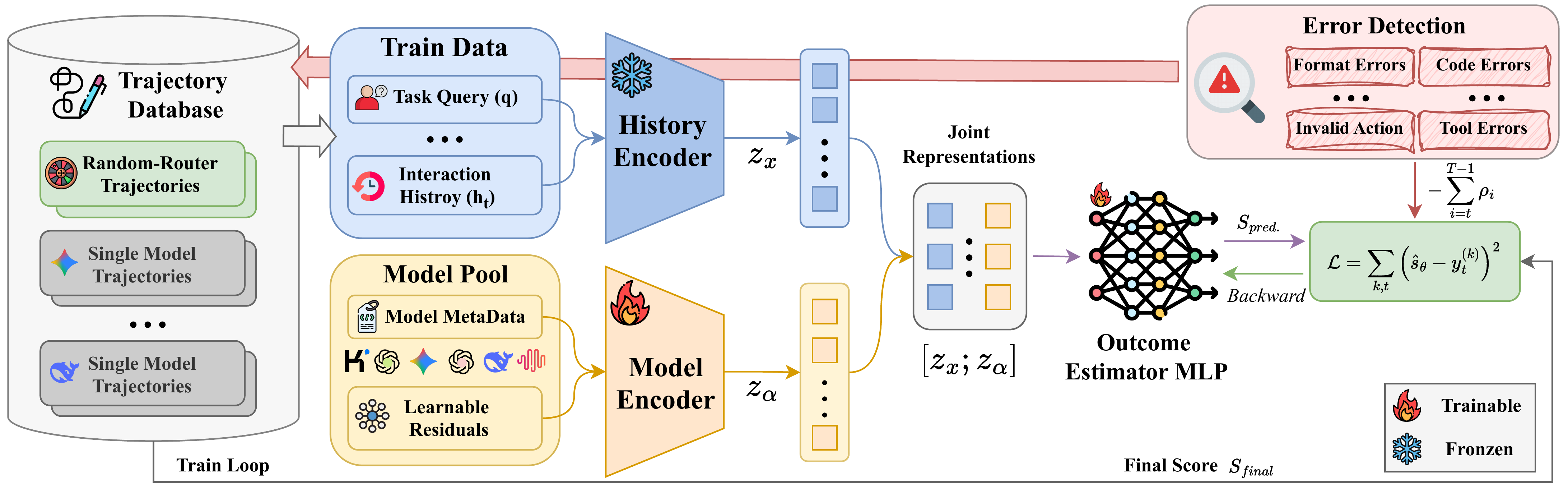}
\caption{\textbf{Overview of \method.}}
\label{fig:method_overview}
\vspace{-1.0em}
\end{figure*}

Single-turn (episode-level) routing makes one model choice at the start of an episode and keeps it fixed. We study \emph{multi-turn} model routing, where the model choice can change over time.
An episode consists of turns indexed by $t$.
At each turn, a router selects a model $a_t \in \mathcal{A}$ to generate the agent's next output (the router itself may be implemented by an LLM).
A \textbf{turn} is one round of interaction that includes exactly one invocation of the selected model: given the \textbf{history} $h_t$ (task description, dialogue context, and the most recent observation), the selected model produces $y_t \sim p_{a_t}(\cdot \mid h_t)$, and a \textbf{parser} maps $y_t$ to an executable action $u_t=\text{parse}(y_t)$; executing $u_t$ yields the next observation $o_{t+1}$.

The episode ends when the task is completed (as indicated by the environment or by an agent submission), a turn limit is reached, or a cost budget is exhausted. The environment provides a terminal score $S_{\text{final}}$.

\paragraph{Objective.}
We optimize task performance under a per-episode cost budget:
\begin{equation}
\max \; \mathbb{E}\left[S_{\text{final}}\right] \quad \text{s.t.} \quad \sum_{t=0}^{T-1} c_t \le B
\label{eq:objective}
\end{equation}
where $c_t$ is the cost at turn $t$ (computed from token usage and model pricing), $B$ is a per-episode cost budget, and $T\le T_{\max}$ is the episode length (capped by a maximum turn limit). Episodes terminate when the budget is exhausted or the turn limit is reached.

\subsection{Joint History--Model Representations}
\label{sec:representations}

Routing decisions depend on both the interaction history and the chosen model (i.e., the pair $(h_t,a_t)$). We therefore embed history and candidate models separately and then combine them into a joint representation used by the router.

\paragraph{History Encoding.}
We represent the routing history as the task block plus the accumulated interaction context. Concretely, we serialize each example with a fixed template:

\begin{equation}
{\textstyle h_t = [q,\langle u_0,o_1\rangle,\ldots,\langle u_{t-1},o_t\rangle]}
\end{equation}
At implementation time, we do not use a fixed number of retained turns K. Instead, we apply token-budget truncation with a maximum length of 8192 tokens: we always keep the task block and retain the most recent interaction context within the remaining budget, truncating the oldest context first. The serialized history is then encoded by a frozen text encoder $\phi$ to obtain $z_x=\phi(h_t)\in\mathbb{R}^d$. 

\paragraph{Model Encoding.}
Each candidate model $a \in \mathcal{A}$ is represented by a learned embedding $z_a=\psi(a)$ that combines (i) a vector of structured attributes $\text{attr}_a$ (including context limits, knowledge cut-off date, and pricing) with (ii) a learned residual $e_a$ that captures model-specific behavior not explained by metadata.The final model embedding concatenates these components through a linear projection:

\begin{equation}
z_a = W_{\text{proj}} \cdot [\text{MLP}(\text{attr}a); e_a] + b{\text{proj}}
\end{equation}
where $e_a$ is the residual embedding regularized to prevent overfitting.



\paragraph{Joint representation.}
We form a joint feature vector by concatenating the two embeddings, $[z_x;z_a]$, and feed it to a shared feed-forward backbone to enable model-conditioned predictions.
To score multiple candidates efficiently, we compute $z_x$ once for the current history and concatenate it with each candidate embedding in a batch.

\subsection{Learning an Outcome Estimator}
\label{sec:learning}

We learn an outcome estimator $\hat{s}_\theta(h_t,a)$ that maps a history--model pair $(h_t,a)$ to the expected terminal outcome when selecting model $a$ at turn $t$.
We parameterize $\hat{s}_\theta$ as a lightweight feed-forward network (an MLP with ReLU nonlinearities and optional dropout) that takes the joint representation $[z_x;z_a]$ and outputs a single scalar:
\begin{equation}
\hat{s}_\theta(h_t,a)=f_\theta([z_x;z_a])\in\mathbb{R}.
\end{equation}
We supervise this estimator with terminal outcomes, rather than dense per-turn rewards, for two reasons: (i) in complex agent environments, faithful intermediate rewards are often unavailable, and (ii) training a stable reward model for dense feedback can be brittle.
In our setting, episodes are executed under both a cost budget $B$ and a maximum turn limit $T_{\max}$, so expensive choices and wasted turns (e.g., errors that trigger retries) are already reflected in the logged terminal scores; we therefore avoid adding a separate cost penalty to the target.

\paragraph{Outcome target with error penalties.}
For each turn in a logged trajectory, we form a turn-conditional target by adjusting the terminal score $S_{\text{final}}$ with a cumulative error penalty.
Concretely, we detect errors (e.g., invalid actions and parsing/format violations) from the trajectory logs and penalize them by severity and turn progress:
\begin{align}
    \label{eq:error}
\tilde{S}_t &= S_{\text{final}} - \sum_{i=t}^{T-1} \rho_i, \\
\rho_i &= \mathbf{1}[\mathcal{E}_i\neq\emptyset]\cdot \beta_{\mathrm{sev}(\mathcal{E}_i)}\cdot w\!\left(\tfrac{i+1}{T_{\max}}\right),
\end{align}
where $\mathcal{E}_i$ is the set of detected error types at turn $i$, $\mathrm{sev}(\mathcal{E}_i)$ takes the maximum severity among errors at that turn, and $\beta$ is the corresponding severity coefficient.
The progress weight $w(\cdot)$ is monotone increasing (we use a simple piecewise-linear schedule), so late errors are penalized more strongly.
This reflects a simple design intuition: early mistakes may be  recoverable as the agent is still gathering information, whereas late mistakes more directly jeopardize task completion, so we impose lower tolerance for errors near the end of an episode.
We train $\hat{s}_\theta(h_t,a)$ to approximate $\mathbb{E}[\tilde{S}_t \mid h_t,a]$.

\paragraph{Training Objective.}
We train on offline trajectories collected under a stochastic router. For each step $(h_t^{(k)}, a_t^{(k)})$ in trajectory $k$, we use the supervision target $y^{(k)}_t=\tilde{S}^{(k)}_t$ and minimize a squared error loss:
\begin{equation}
\mathcal{L} = \sum_{k,t} \left( \hat{s}_\theta(h_t^{(k)}, a_t^{(k)}) - y^{(k)}_t \right)^2
\end{equation}

\paragraph{Data Collection.}
We log each episode as a trajectory (a sequence of turn-level tuples) and collect training trajectories from two sources:
(i) a \textbf{uniform (random) router} that samples models from the pool at each turn, and
(ii) \textbf{single-model} runs where one candidate model is used for the entire episode on training tasks.
The random-router data provides broad coverage of model choices, while the single-model data anchors the estimator with consistent per-model behavior.
Across both benchmarks, the offline training set contains 1,291 training instances, 29,693 trajectories, and 515,221 total turns, with an estimated one-time collection cost of approximately \$1,620.

\paragraph{Inference.}
At deployment, the router selects the model with the highest predicted outcome,
\begin{equation}
a_t^* = \arg\max_{a \in \mathcal{A}} \hat{s}_\theta(h_t, a),
\end{equation}
and the episode is executed under the per-episode budget $B$ and turn limit $T_{\max}$ (terminating when either limit is reached).

\section{Experiments}
\label{sec:experiments}

We evaluate \method on two challenging multi-turn benchmarks and analyze the learned routing patterns.
Unless otherwise stated, we repeat each experiment three times and report the mean.
Our anonymous code repository is available at \github.

\subsection{Experimental Setup}

\paragraph{Benchmarks.}
We evaluate on two multi-turn benchmarks: \textbf{ScienceWorld} \citep{wang2022scienceworld} and \textbf{HLE (Humanity's Last Exam)}~\citep{phan2025humanitysexam}. 
ScienceWorld is a text-based interactive environment requiring procedural scientific reasoning, with a terminal score $S_{\text{final}} \in [-100, 100]$.
HLE is a long-context benchmark spanning academic domains, where questions require multi-step reasoning with tool use and success is binary.
We consider both in-distribution (ID) and out-of-distribution (OOD) splits.
We construct OOD evaluations to be \emph{semantically} disjoint from training and ID test (no overlap), rather than relying on random re-sampling, by holding out entire task types / subject categories as OOD.
For ScienceWorld, we use 13 task types for ID and reserve 12 held-out task types for OOD; the full task-type split is listed in Table~\ref{tab:sw_tasks}; we further split task \emph{variations} within the ID task types into 60\%/20\%/20\% train/validation/test (Appendix~\ref{app:data}). 
For HLE, we use 6 subject categories for ID and hold out 2 categories for OOD; the detailed category split is listed in Table~\ref{tab:hle_categories}; 
we report ID vs.\ OOD performance by partitioning the benchmark's test questions into the chosen subject groups (Table~\ref{tab:hle_categories}).

\paragraph{Tool Configuration.}
For HLE, we follow the tool-use setting of \texttt{TongYi-DeepResearch}~\citep{team2025tongyi} and enable four tools: \texttt{search}, \texttt{browse}, \texttt{python}, and \texttt{answer}.
\texttt{search} uses Serper's Google Search API, while \texttt{browse} fetches webpage content (via Jina Reader when enabled) and optionally summarizes long pages to keep the context bounded (Appendix~\ref{app:prompts:browse}); \texttt{python} supports deterministic computation and \texttt{answer} submits the final response.
ScienceWorld does not require external web tools; the agent interacts with the simulator via a single text-action command per turn, with built-in query commands (e.g., \texttt{?navigation}, \texttt{?object}) to enumerate valid actions.
Tool schemas are injected into the HLE system prompt at runtime and are listed in Appendix~\ref{app:tool_schemas}.

\paragraph{Error Detection.}
We detect errors from environment observations to compute annealed error costs during training.
Table~\ref{tab:error_categories} summarizes the error categories by benchmark.
HLE errors span format violations, Python execution failures, and tool-specific issues; ScienceWorld only penalizes unparseable actions (environmental feedback like ``door is not open'' is normal exploration, not an error).
Severity levels (high/medium/low) determine penalty coefficients in the AEC computation.
Full rule specifications are provided in Appendix~\ref{app:error_rules}.
Unless otherwise stated, we use severity coefficients high=1.0, medium=0.8, and low=0.2 throughout.

\begin{table}[t]
\centering
\small
\begin{tabular}{llc}
\toprule
\textbf{Benchmark} & \textbf{Error Category} & \textbf{\# Rules} \\
\midrule
\multirow{4}{*}{HLE} & Format Errors & 4 \\
 & Python Execution Errors & 11 \\
 & Search Tool Errors & 3 \\
 & Browse Tool Errors & 5 \\
\midrule
ScienceWorld & Invalid Action & 1 \\
\bottomrule
\end{tabular}
\caption{Error categories used for error detection. HLE has diverse tool-related errors, while ScienceWorld only penalizes unparseable actions. Full rule specifications are in Appendix~\ref{app:error_rules}.}
\label{tab:error_categories}
\vspace{-1.0em}
\end{table}

\paragraph{Model Pool.}
We evaluate with 6 frontier LLMs spanning a 20$\times$ cost range (see in Table~\ref{tab:model_pool}).

\begin{table}[h]
\centering
\small
\resizebox{\linewidth}{!}{
\begin{tabular}{lrrr}
\toprule
\textbf{Model} & \textbf{Context} & \textbf{In \$/M} & \textbf{Out \$/M} \\
\midrule
GPT-5~\citep{openai2025gpt5api} & 400K & 1.25 & 10.00 \\
DeepSeek-V3.2~\citep{liu2025deepseek} & 164K & 0.27 & 0.42 \\
MiniMax-M2~\citep{minimax2025m2} & 197K & 0.20 & 1.00 \\
Kimi-K2~\citep{team2025kimi} & 131K & 0.39 & 1.90 \\
Gemini-2.5-Flash-Lite~\citep{kilpatrick2025gemini25flashlite} & 1M & 0.10 & 0.40 \\
GPT-OSS-120B~\citep{openai2025gptoss120bgptoss20bmodel} & 131K & 0.09 & 0.36 \\
\bottomrule
\end{tabular}
}
\caption{Model pool with pricing (from OpenRouter).}
\label{tab:model_pool}
\vspace{-1.0em}
\end{table}

\paragraph{Implement Details.}
We train for 100 epochs with early stopping (patience=3) using AdamW optimizer (lr=$10^{-3}$, weight decay=0.01) and cosine annealing. Batch size is 64.
We encode histories with a frozen \texttt{Qwen/Qwen3-Embedding-0.6B} encoder that produces 1024-dimensional embeddings, and we set the maximum input length to 8192 tokens.
The model encoder maps an 8-d metadata feature vector to a 32-d attribute embedding, concatenates it with a 16-d per-model residual embedding, and linearly projects the resulting 48-d vector to a 64-d model embedding. To prevent premature convergence, we apply an $L_2$ penalty ($\lambda=0.001$) to the learnable residual embeddings.
For the error penalty in Eq.~\ref{eq:error}, we instantiate the progress weight $w(\cdot)$ as a piecewise-linear warmup with $p_0{=}0.3$, $p_1{=}0.7$, $w_{\min}{=}0.3$, and $w_{\max}{=}1.0$ (Appendix~\ref{app:error_rules}).
During evaluation, we enforce a maximum horizon of 50 steps for ScienceWorld and 30 steps for HLE, with a per-episode cost cap of \$2.0 for both benchmarks.

\begin{table*}[!htbp]
\centering
\small
\setlength{\tabcolsep}{5pt}
\renewcommand{\arraystretch}{1.08}
\resizebox{\linewidth}{!}{%
\begin{tabular}{l cc cc cc cc}
\toprule
& \multicolumn{4}{c}{\textbf{ScienceWorld}} & \multicolumn{4}{c}{\textbf{HLE}} \\
\cmidrule(lr){2-5}\cmidrule(lr){6-9}
& \multicolumn{2}{c}{\textit{Test}} & \multicolumn{2}{c}{\textit{OOD}} & \multicolumn{2}{c}{\textit{Test}} & \multicolumn{2}{c}{\textit{OOD}} \\
\cmidrule(lr){2-3}\cmidrule(lr){4-5}\cmidrule(lr){6-7}\cmidrule(lr){8-9}
\multirow{-3}{*}{\textbf{Method}} & Score$\textcolor{green!60!black}{\uparrow}$ & Total Cost (\$)$\textcolor{red!70!black}{\downarrow}$ & Score$\textcolor{green!60!black}{\uparrow}$ & Total Cost (\$)$\textcolor{red!70!black}{\downarrow}$ & Acc$\textcolor{green!60!black}{\uparrow}$ & Total Cost (\$)$\textcolor{red!70!black}{\downarrow}$ & Acc$\textcolor{green!60!black}{\uparrow}$ & Total Cost (\$)$\textcolor{red!70!black}{\downarrow}$ \\
\midrule
\multicolumn{9}{c}{\textit{Single-Model Baselines}} \\
\midrule
GPT-5 & 48.4$\pm$2.1 & 13.9 & 4.9$\pm$2.9 & 47.6 & 25.1$\pm$1.6 & 61.8 & 34.8$\pm$2.2 & 65.3 \\
DeepSeek-V3.2 & 13.1$\pm$2.0 & 2.9 & -4.2$\pm$2.4 & 22.8 & 15.6$\pm$1.3 & 22.4 & 28.7$\pm$1.6 & 22.2 \\
MiniMax-M2 & -0.5$\pm$1.6 & 3.2 & 0.9$\pm$2.3 & 10.9 & 7.8$\pm$1.4 & 18.1 & 8.9$\pm$1.9 & 9.8 \\
Kimi-K2 & 5.2$\pm$1.8 & 2.5 & 0.2$\pm$2.1 & 8.9 & 11.4$\pm$1.2 & 12.0 & 20.1$\pm$1.5 & 9.8 \\
Gemini-2.5-Flash-Lite & 4.2$\pm$1.7 & 0.3 & -2.1$\pm$2.2 & 1.5 & 5.6$\pm$1.1 & 3.0 & 8.4$\pm$1.5 & 2.1 \\
GPT-OSS-120B & 26.6$\pm$2.3 & 0.5 & 1.1$\pm$2.7 & 4.2 & 9.7$\pm$1.0 & 0.7 & 11.4$\pm$1.2 & 2.1 \\
\midrule
\multicolumn{9}{c}{\textit{Single-Turn Routers (episode-level)}} \\
\midrule
RouterDC \citep{chen2024routerdc} & 23.1$\pm$2.4 & 3.3 & 5.5$\pm$3.1 & 2.5 & 12.8$\pm$1.6 & 10.3 & 17.9$\pm$2.1 & 13.5 \\
EmbedLLM \citep{wang2024embedllm} & 30.4$\pm$2.8 & 5.6 & 5.0$\pm$3.4 & 3.0 & 24.8$\pm$2.0 & 61.4 & 33.6$\pm$2.5 & 56.8 \\
AvengersPro \citep{zhang2025avengers} & 36.8$\pm$2.5 & 4.1 & 2.4$\pm$3.2 & 4.1 & 23.7$\pm$1.8 & 47.5 & 30.6$\pm$2.3 & 33.3 \\
\midrule
\multicolumn{9}{c}{\textit{Multi-Turn Routers (turn-level)}} \\
\midrule
Random Router & 21.7$\pm$4.5 & 3.9 & -8.1$\pm$5.4 & 20.3 & 20.0$\pm$2.8 & 16.9 & 23.8$\pm$3.4 & 14.7 \\
LLM Router & 19.8$\pm$3.8 & 12.2 & -0.4$\pm$4.6 & 28.3 & 24.0$\pm$2.3 & 56.2 & 36.0$\pm$2.8 & 35.6 \\
Router-R1~\citep{zhang2025router} & 42.1$\pm$3.6 & 12.6 & 2.1$\pm$4.2 & 21.0 & 24.2$\pm$2.2 & 51.9 & 35.1$\pm$2.6 & 60.7 \\
OpenRouter$^{\dagger}$ \citep{openrouter} & -26.4$\pm$3.2 & 3.0 & -26.9$\pm$3.8 & 15.5 & 18.3$\pm$2.1 & 138.5 & 34.0$\pm$2.5 & 154.3 \\
\midrule
\rowcolor{blue!10}
\textbf{\method~(ours)} & \textbf{53.8$\pm$3.2} & 5.7 & \textbf{9.9$\pm$3.9} & 16.3 & \textbf{26.0$\pm$2.3} & 35.0 & \textbf{38.6$\pm$3.0} & 31.2 \\
\rowcolor{blue!6}
\multicolumn{1}{l}{\quad\scriptsize\textit{$\Delta$ vs GPT-5}} &
{\scriptsize\textcolor{blue!70!black}{\textbf{+5.4}}} &
{\scriptsize\textcolor{orange!85!black}{\textit{saving 58.7\%}}} &
{\scriptsize\textcolor{blue!70!black}{\textbf{+5.0}}} &
{\scriptsize\textcolor{orange!85!black}{\textit{saving 65.8\%}}} &
{\scriptsize\textcolor{blue!70!black}{\textbf{+0.9}}} &
{\scriptsize\textcolor{orange!85!black}{\textit{saving 43.4\%}}} &
{\scriptsize\textcolor{blue!70!black}{\textbf{+3.8}}} &
{\scriptsize\textcolor{orange!85!black}{\textit{saving 52.3\%}}} \\
\rowcolor{blue!6}
\multicolumn{1}{l}{\quad\scriptsize\textit{$\Delta$ vs Router-R1}} &
{\scriptsize\textcolor{blue!70!black}{\textbf{+11.7}}} &
{\scriptsize\textcolor{orange!85!black}{\textit{saving 54.4\%}}} &
{\scriptsize\textcolor{blue!70!black}{\textbf{+7.8}}} &
{\scriptsize\textcolor{orange!85!black}{\textit{saving 22.4\%}}} &
{\scriptsize\textcolor{blue!70!black}{\textbf{+1.8}}} &
{\scriptsize\textcolor{orange!85!black}{\textit{saving 32.7\%}}} &
{\scriptsize\textcolor{blue!70!black}{\textbf{+3.5}}} &
{\scriptsize\textcolor{orange!85!black}{\textit{saving 48.7\%}}} \\
\bottomrule
\end{tabular}
}
\caption{Main results on ScienceWorld and HLE benchmarks (Test and OOD splits). We report mean$\pm$std over three runs for the performance metrics; total cost is summed over evaluated episodes. OOD evaluations use held-out task types (ScienceWorld) and held-out subject categories (HLE). The $\Delta$ rows show score gains and relative cost savings compared to GPT-5 and Router-R1. $^{\dagger}$OpenRouter uses a fixed provider-side routing API with a different model pool (Appendix~\ref{app:openrouter}).}
\label{tab:combined_results}
\end{table*}

\paragraph{Baselines.}
We compare against three groups of baselines:
\begin{itemize}[nolistsep, noitemsep, leftmargin=*]
\item \textbf{Single-model}: each candidate model from our pool is used exclusively for the entire episode.
\item \textbf{Single-turn routers (episode-level)}: \textbf{RouterDC}~\citep{chen2024routerdc}, \textbf{EmbedLLM}~\citep{wang2024embedllm}, and \textbf{AvengersPro}~\citep{zhang2025avengers, zhang2025beyond}.
These routers make a single routing decision at the start of an episode based on the initial query context, then keep the selected model fixed for all subsequent turns.
\item \textbf{Multi-turn routers (turn-level)}: \textbf{Random Router} (uniform random selection at each turn), \textbf{Router-R1}~\citep{zhang2025router} (we train a \texttt{Qwen2.5-7B-Instruct} routing model following the Router-R1 recipe), \textbf{LLM Router} (same prompt as Router-R1 but directly using \texttt{DeepSeek-V3.2} as the routing model, no training), and \textbf{OpenRouter}~\citep{openrouter} (a representative commercial router via OpenRouter's automatic routing API).
OpenRouter is included to contextualize \method against an off-the-shelf production routing system; however, its routing API does not allow us to customize the candidate model pool, and it  selects from a substantially larger pool than our fixed 6-model setting (Appendix~\ref{app:openrouter}).
\end{itemize}

\subsection{Does MTRouter Work?}
\label{sec:main_results}
Table~\ref{tab:combined_results} presents in-distribution (test) and out-of-distribution (OOD) results across MTRouter and the other baselines. 
\paragraph{Test.}
On ScienceWorld, \method achieves the best average score (53.8) while cutting total cost by 58.7\% vs.\ GPT-5; compared to Router-R1, it gains +11.7 points with 54.4\% lower total cost.
Episode-level routers (single-turn) that commit to one model per episode consistently lag behind, supporting the necessity of \emph{multi-turn} routing in interactive settings where phases and errors evolve over time.
Notably, OpenRouter produces negative scores on ScienceWorld because it underestimates task difficulty and over-relies on lightweight models (Appendix~\ref{app:openrouter}).
On HLE, MTRouter attains the best accuracy (26.0\%) while remaining cost-efficient (43.4\% lower total cost than GPT-5 and 32.7\% lower than Router-R1); Router-R1 and LLM Router reach similar accuracy but at higher cost. 
Overall, \method delivers a better accuracy--cost trade-off than both strong single-model baselines and existing routing baselines on both benchmarks.

\begin{table}[!htbp]
\centering
\small
\setlength{\tabcolsep}{5pt}
\renewcommand{\arraystretch}{1.06}
\resizebox{\linewidth}{!}{%
\begin{tabular}{lcc}
\toprule
\textbf{Variant} & \textbf{SW Avg Score}$\textcolor{green!60!black}{\uparrow}$ & \textbf{HLE Acc. (\%)}$\textcolor{green!60!black}{\uparrow}$ \\
\midrule
\rowcolor{blue!10}\textbf{\method (full setting)} & 53.8 $\pm$ 3.2 & 26.0 $\pm$ 2.3 \\
\midrule
Ridge instead of MLP & 49.1 $\pm$ 3.5 & 23.4 $\pm$ 2.2 \\
w/o Random-Router data & 47.2 $\pm$ 4.1 & 22.6 $\pm$ 2.1 \\
w/o error penalties & 48.5 $\pm$ 3.6 & 23.8 $\pm$ 2.1 \\
w/o routing history$^{\ast}$ & 44.6 $\pm$ 3.8 & 21.3 $\pm$ 2.0 \\
Hardcoded model encoder & 41.3 $\pm$ 4.4 & 19.7 $\pm$ 2.1 \\
\bottomrule
\end{tabular}
}
\caption{Ablation study on ScienceWorld and HLE test sets. We report performance with 95\% confidence intervals. $^{\ast}$This ablation removes \emph{router} history: routing conditions only on the current turn (the chosen model still receives the full conversation history).}
\label{tab:ablation}
\vspace{-1.0em}
\end{table}

\paragraph{OOD.}
We next examine the OOD columns of Table~\ref{tab:combined_results}, which evaluate semantic distribution shift (held-out ScienceWorld task types and held-out HLE subject categories).
On ScienceWorld OOD, \method improves over GPT-5 by +5.0 points while using 65.8\% lower total cost; on HLE OOD, it reaches 38.57\% accuracy with 52.3\% lower total cost than GPT-5.
These results show that \method not only generalizes under distribution shift but also preserves its cost efficiency, establishing the strongest overall performance among the compared methods.

\paragraph{Ablation Studies}

Table~\ref{tab:ablation} summarizes ablations on ScienceWorld and HLE.
Performance degrades when we replace the MLP with a simpler regressor, remove random-router data, or remove the error-penalty adjustment, indicating that each component contributes to learning reliable routing preferences from offline trajectories.
The largest drops come from removing routing history or replacing the learned model encoder with hardcoded features, highlighting the importance of modeling both the evolving interaction context and model behavior.
We use Ridge as a standard regularized linear baseline to test whether the gains come from nonlinear modeling rather than feature construction alone.
Additional robustness checks on budget sensitivity, candidate-pool size, and history token budget are reported in Appendix~\ref{app:more_experiments}.

\begin{figure}[!htbp]
\centering
\includegraphics[width=\columnwidth]{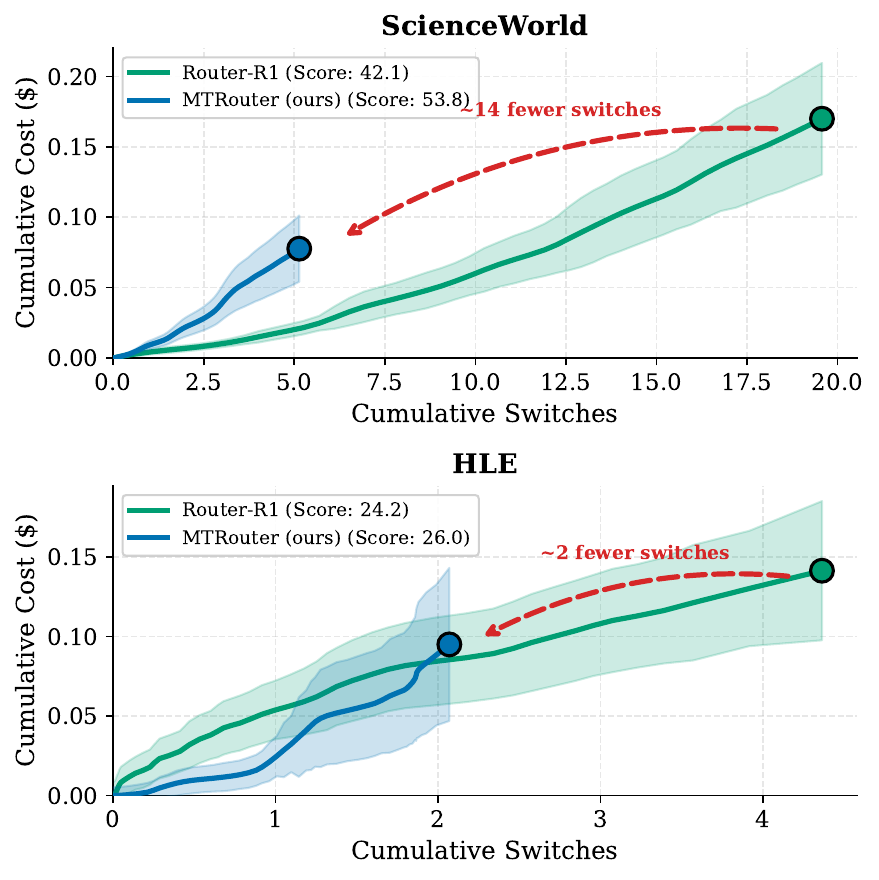}
\caption{Cumulative cost vs.\ cumulative model switches over successful episodes, comparing \method against Router-R1 on ScienceWorld and HLE (constructed by replaying logged trajectories).}
\label{fig:cost_switches}
\vspace{-1.0em}
\end{figure}

\subsection{Why Does \method Work?}
\label{sec:analysis}
While the in-domain and out-of-domain results (Table~\ref{tab:combined_results}) and ablations (Table~\ref{tab:ablation}) establish \method's effectiveness, we next ask a more diagnostic question: \emph{why} does it work?
We use a sequence of complementary analyses to connect the performance gains to concrete routing behaviors.

\begin{figure}[!htbp]
\centering
\includegraphics[width=\columnwidth]{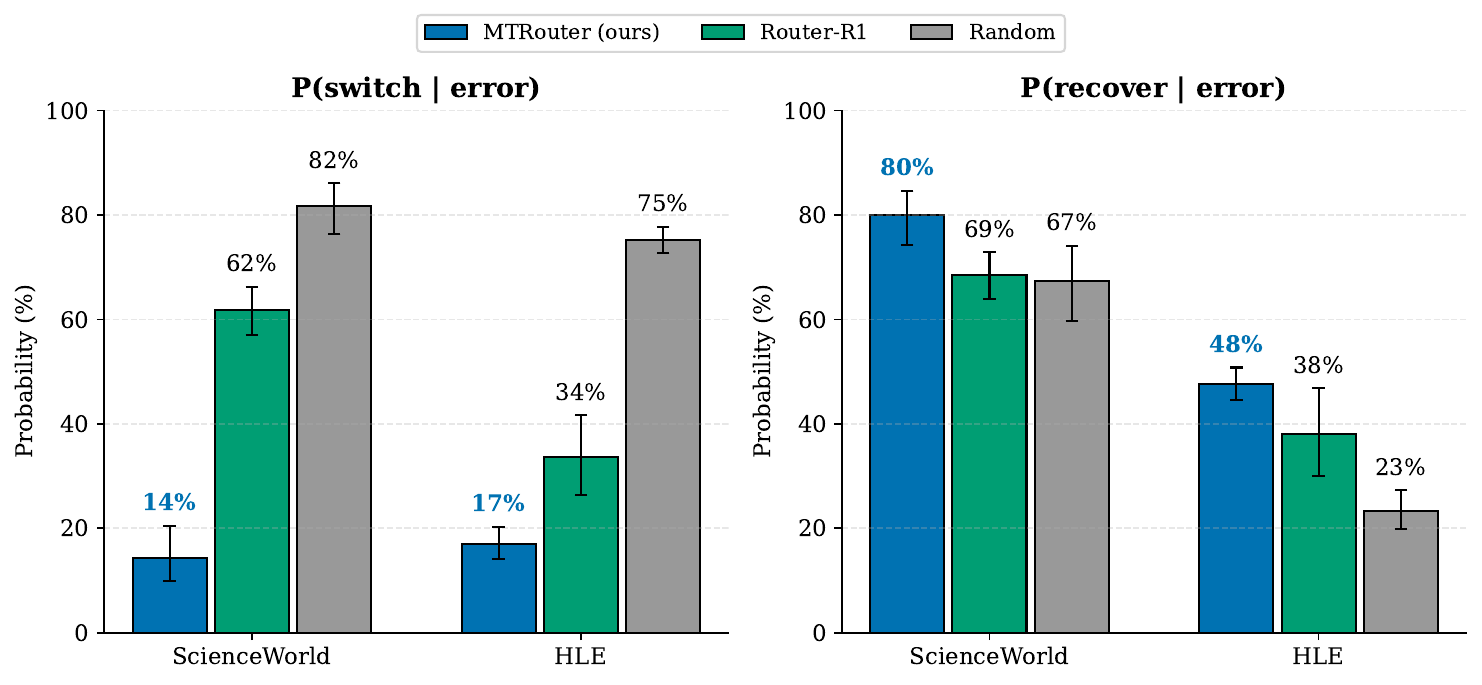}
\caption{Error-triggered switching and recovery. \textbf{Left}: probability of switching models after an error (format errors / invalid actions). \textbf{Right}: probability that the next turn recovers (no error) conditioned on an error.}
\label{fig:error_switch_recovery}
\vspace{-1.0em}
\end{figure}

\paragraph{Start from a simple diagnostic: switching vs.\ cost.}
If multi-turn routing is ``just switch more often,'' then a router that switches frequently should dominate.
We find the opposite.
Figure~\ref{fig:cost_switches} plots, over \emph{successful} episodes, how cumulative API cost grows as the router makes additional model switches.
Each curve is constructed by replaying logged trajectories from \method and Router-R1, accumulating per-turn cost and counting switches along the episode.
Despite \method achieving better end performance (Table~\ref{tab:combined_results}), its trajectories typically reach success with \emph{fewer} switches and \emph{lower} cumulative cost (e.g., on ScienceWorld: $\sim$5 switches for \method vs.\ $\sim$20 for Router-R1).
\begin{figure}[!htbp]
\centering
\includegraphics[width=\linewidth]{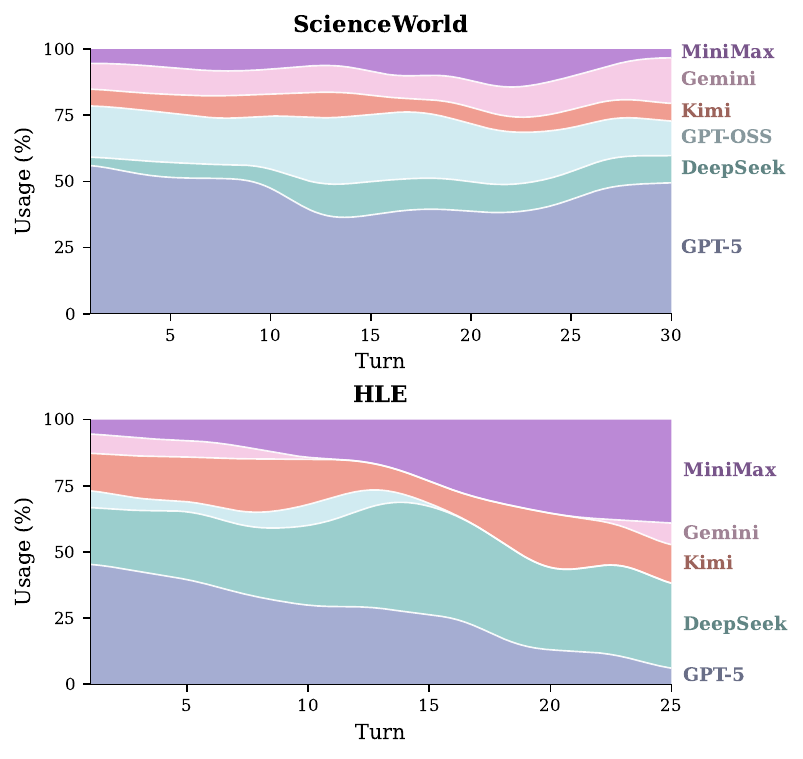}
\caption{Model usage by turn on ScienceWorld and HLE. \method exhibits structured, benchmark-specific routing behavior rather than uniformly switching models across turns. In the HLE, GPT-OSS-120B is used primarily in early turns and then decays to near-zero usage in later turns, which makes its band difficult to notice visually.}
\label{fig:phase_usage}
\vspace{-1.0em}
\end{figure}

\begin{figure*}[!htbp]
\centering
\includegraphics[width=\linewidth]{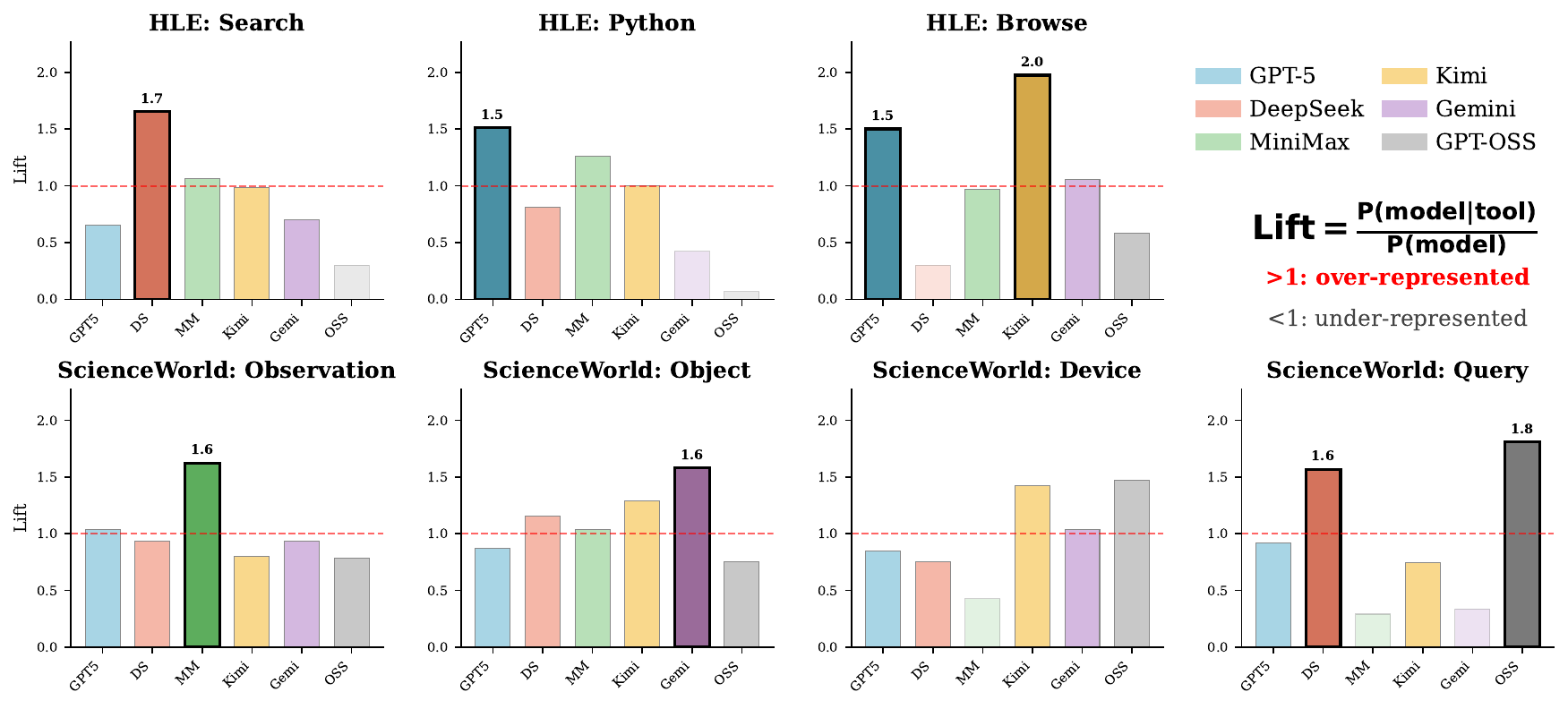}
\caption{Tool/action specialization by model measured via \emph{lift} ($\mathrm{Lift}=P(\text{model}\mid \text{tool})/P(\text{model})$). Values $>1$ indicate a model is over-represented for a tool/action (specialization), while values $<1$ indicate under-use.}
\label{fig:action_by_model}
\vspace{-1.0em}
\end{figure*}

Beyond differences in per-token pricing, frequent switching can also reduce the effectiveness of prompt caching in multi-turn settings, lowering cache hit rates and increasing the effective cost of serving long histories \citep{hu2025hands}.
This immediately raises a more specific question: \emph{when} does Router-R1 choose to switch, and are those switches actually helpful?

\paragraph{Switching less by being more tolerant to errors.}
Figure~\ref{fig:error_switch_recovery} shows that Router-R1 switches aggressively after errors, while \method is less reactive: it often keeps the current model and tries to continue.
Crucially, this is not ``ignoring'' errors: the right panel shows a higher probability of recovery on the next turn under \method.
Consistent with this, after an error \method stays with the same model $\approx$90.2\% of the time on ScienceWorld and $\approx$80.9\% on HLE, substantially higher than Router-R1 (38.3\% and 66.4\%, respectively).
Together with Figure~\ref{fig:cost_switches}, these trends suggest that \method avoids a large fraction of error-triggered switches that appear low-yield, helping control cumulative cost without sacrificing performance.
We hypothesize this gap stems from the learning signal: Router-R1 largely relies on a natural-language router prompt to infer when switching helps, whereas \method is trained directly on trajectory outcomes (terminal scores with annealed error penalties), providing more direct supervision for effective switching.

\paragraph{Not ``never switch''---but switch with structure.}
One might worry that the previous results simply reflect a conservative router that rarely changes models.
Figure~\ref{fig:phase_usage} rules this out: \method uses multiple models throughout an episode, but in a stable, benchmark-specific way rather than as a reflex to errors.
This suggests that the router is learning a \emph{strategy} (which models to rely on, and when), not just a generic ``upgrade on failure'' heuristic.
For instance, on ScienceWorld, GPT-5 accounts for 50.8\% of early turns, while GPT-OSS increases to 24.3\% in the final turns; in contrast, Router-R1 largely concentrates on DeepSeek and Gemini at roughly $\sim$45\% each across phases, exhibiting much less structured diversity.

\paragraph{A concrete form of strategy: emergent specialization.}
To make this structure explicit, Figure~\ref{fig:action_by_model} measures \emph{lift}: how much a model is over-used for a tool/action relative to its overall frequency.
We observe clear specialization patterns (lift $>1$) that align with complementary strengths---e.g., on HLE, DeepSeek is over-represented on \texttt{search} (lift 1.66), GPT-5 on \texttt{python} (1.51), and Kimi on \texttt{browse} (1.98).
On ScienceWorld, we observe analogous specialization across action types, such as MiniMax on observation-heavy actions (1.62), Gemini on object interactions (1.58), and GPT-OSS on query commands (1.81).
These findings connect back to the main results: \method wins not by switching more, but by switching \emph{selectively} and assigning stable roles to models over the course of an episode.

\begin{figure}[t]
\centering
\includegraphics[width=\columnwidth]{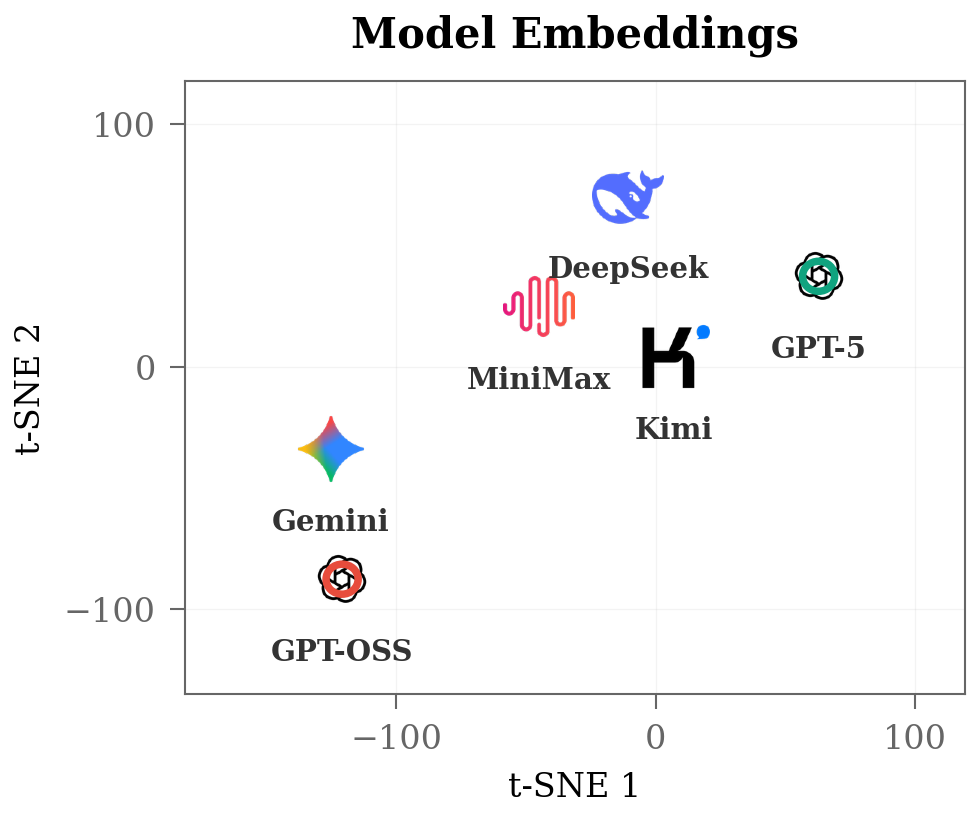}
\caption{t-SNE visualization of learned model embeddings from the model encoder. The embeddings separate models by identity and form a clear cost-tier structure, with low-cost models (e.g., GPT-OSS, Gemini) distinct from higher-cost frontier models (e.g., GPT-5).}
\label{fig:model_embeddings}
\end{figure}

\paragraph{Learned model embeddings.}

Figure~\ref{fig:model_embeddings} visualizes the learned model embeddings after training.
The encoder learns to distinguish the candidate models and organizes them by cost tier, suggesting it captures meaningful capability--cost structure beyond raw attributes.

\section{Conclusion}

In this paper, we presented \method, a multi-turn routing framework that optimizes the performance--cost trade-off in agentic workflows via an offline-learned outcome estimator. By enabling turn-level model selection, our approach allows agents to adaptively allocate computational resources based on the evolving interaction state.
Our experiments on ScienceWorld and HLE demonstrate the empirical effectiveness of \method. It improves the average score on ScienceWorld from 48.4 to 53.8 while reducing total costs by 58.7\%. On HLE, it achieves a 43.4\% cost reduction while maintaining competitive accuracy. Beyond aggregate performance, our analysis of the routing trajectories shows that \method exhibits structured model usage across different actions and achieves success with fewer model switches than existing baselines.

\section*{Limitations}

Despite the performance and cost gains demonstrated by \method, several limitations remain that offer avenues for future work.

Scalability of Data Collection. The primary bottleneck for training turn-level routers is the cost of collecting diverse, long-horizon trajectories across multiple candidate models. Due to computational budget constraints, we evaluate episodes with horizons of up to 50 steps on ScienceWorld and 30 steps on HLE.

Lack of Online Adaptation. Currently, \method operates in an offline learning paradigm. While this significantly reduces the cost of training, the router cannot adapt its strategy in real-time to novel or rapidly evolving environments. Implementing online updates via reinforcement learning would likely yield further gains in robustness, but the prohibitive cost of continuous online interaction with frontier models makes this challenging for current academic research.

Prompt Caching and Switching Overhead. A technical challenge inherent to multi-turn routing is the potential loss of prompt caching (e.g., KV cache) when switching between different models. In many API-based deployments, frequent switching requires the new model to re-process the entire trajectory prefix, which could inadvertently increase latency and per-turn costs. However, our analysis suggests that \method partially mitigates this issue; unlike reactive baselines that switch models at the first sign of a transient error, \method exhibits a more "composed" switching pattern, maintaining model stability over longer sequences of turns. This structured model usage, as reflected in our cost-efficiency results, helps balance the trade-off between adaptive model selection and the benefits of context caching.

\section*{Acknowledgements}
The work was supported by the National Natural Science Foundation of China (Nos. 62272092, 62172086, 62506186), and the Fundamental Research Funds for the Central Universities under Grants (N25XQD004). This work was supported by Shanghai Artificial Intelligence Laboratory. This work was done during Yiqun Zhang, Hao Li's internships at Shanghai Artificial Intelligence Laboratory.

\bibliography{custom}

\appendix

\section{Dataset and Split Details}
\label{app:data}

\subsection{ScienceWorld Task Types}

Table~\ref{tab:sw_tasks} lists the ScienceWorld task types used for training and out-of-distribution evaluation.

\begin{table}[h]
\centering
\small
\resizebox{\linewidth}{!}{%
\begin{tabular}{ll}
\toprule
\textbf{Split} & \textbf{Task Types} \\
\midrule
\multirow{4}{*}{Train (13)} & boil, melt, chemistry-mix, find-animal, \\
& find-plant, grow-plant, identify-life-stages-1, \\
& lifespan-longest-lived, inclined-plane-angle, \\
& measure-melting-point, power-component, \\
& test-conductivity, mendelian-genetics \\
\midrule
\multirow{4}{*}{OOD Test (12)} & freeze, change-state-of-matter, \\
& chemistry-mix-paint, find-non-living, \\
& grow-fruit, identify-life-stages-2, \\
& lifespan-shortest, inclined-plane-friction, \\
& use-thermometer, power-renewable, \\
& test-conductivity-unknown, genetics-unknown \\
\bottomrule
\end{tabular}
}
\caption{ScienceWorld task type splits.}
\label{tab:sw_tasks}
\end{table}

For each task type, we sample up to 30 variations (to bound collection time). Variations are split 60\%/20\%/20\% into train/validation/test using a fixed random seed (42).

\subsection{HLE Subject Categories}

\begin{table}[h]
\centering
\small
\begin{tabular}{lc}
\toprule
\textbf{Category} & \textbf{N (Test)} \\
\midrule
\multicolumn{2}{c}{\textit{In-Distribution (Training)}} \\
\midrule
Math & 200 \\
Physics & 46 \\
Chemistry & 23 \\
Biology/Medicine & 37 \\
Engineering & 16 \\
Computer Science/AI & 37 \\
\midrule
\multicolumn{2}{c}{\textit{Out-of-Distribution}} \\
\midrule
Humanities/Social Science & 193 \\
Other & 176 \\
\bottomrule
\end{tabular}
\caption{HLE test set distribution by category.}
\label{tab:hle_categories}
\end{table}

\section{Error Detection Rules}
\label{app:error_rules}

Table~\ref{tab:error_rules_detail} provides the full specification of error detection rules used to compute annealed error costs (AEC) during training.
Each rule consists of pattern strings matched against environment observations, a severity level (high/medium/low), and a description.

\paragraph{Design Principles.}
For HLE, we distinguish between model errors (format violations, Python exceptions) and external failures (HTTP errors, paywalls).
Model errors receive higher severity since they reflect controllable mistakes; external failures are marked low severity as they depend on third-party services.
For ScienceWorld, we only penalize truly invalid actions (commands not recognized by the environment parser).
Environmental feedback such as ``the door is not open'' or ``the object is already in your inventory'' represents normal exploration and is not treated as an error.

\paragraph{Severity Coefficients.}
Severity levels map to penalty coefficients in AEC: high (1.0), medium (0.8), low (0.2).
The warmup schedule uses $p_0{=}0.3$, $p_1{=}0.7$, $w_{\min}{=}0.3$, and $w_{\max}{=}1.0$.
These coefficients modulate the base penalty $1/N$ where $N$ is the expected episode length for the task category.

\begin{table*}[!htbp]
\centering
\small
\begin{tabular}{lllp{7cm}}
\toprule
\textbf{Category} & \textbf{Rule Name} & \textbf{Severity} & \textbf{Description} \\
\midrule
\multicolumn{4}{l}{\textit{\textbf{HLE Benchmark}}} \\
\midrule
\multirow{4}{*}{Format Errors}
 & format\_error & medium & Tool call format error---model did not follow required format \\
 & tool\_invalid\_args & medium & Invalid tool arguments or missing required parameters \\
 & tool\_parse\_error & medium & Tool call parsing failure \\
 & tool\_unknown & high & Model called a non-existent tool name \\
\midrule
\multirow{11}{*}{\shortstack[l]{Python\\Execution\\Errors}}
 & python\_traceback & high & Python execution exception with traceback \\
 & python\_name\_error & high & Undefined variable reference \\
 & python\_syntax\_error & high & Python syntax error \\
 & python\_indentation\_error & high & Python indentation error \\
 & python\_type\_error & high & Type mismatch error \\
 & python\_value\_error & medium & Invalid value error \\
 & python\_index\_error & medium & Index out of bounds \\
 & python\_key\_error & medium & Dictionary key not found \\
 & python\_attribute\_error & medium & Attribute access error \\
 & python\_import\_error & medium & Module import failure \\
 & python\_zero\_division & medium & Division by zero \\
 & python\_timeout & high & Code execution timeout \\
\midrule
\multirow{3}{*}{\shortstack[l]{Search Tool\\Errors}}
 & search\_no\_results & high & Search returned no results \\
 & search\_http\_error & low & HTTP connection error during search \\
 & search\_rate\_limit & low & Search rate limit exceeded \\
\midrule
\multirow{5}{*}{\shortstack[l]{Browse Tool\\Errors}}
 & browse\_403 & low & HTTP 403 Forbidden \\
 & browse\_404 & low & HTTP 404 Not Found \\
 & browse\_access\_denied & low & Access denied to resource \\
 & browse\_paywall & low & Paywall or subscription block \\
 & browse\_timeout & low & Browse request timeout \\
\midrule
\multicolumn{4}{l}{\textit{\textbf{ScienceWorld Benchmark}}} \\
\midrule
Invalid Action & no\_known\_action & high & Invalid action command not recognized by environment \\
\bottomrule
\end{tabular}
\caption{Detailed error detection rules used for annealed error cost (AEC) computation. Severity levels determine penalty coefficients: high (1.0), medium (0.8), low (0.2). Browse tool errors are marked low severity as they reflect external failures rather than model errors.}
\label{tab:error_rules_detail}
\end{table*}

\section{Additional Experiments}
\label{app:more_experiments}

\begin{table}[!htbp]
\centering
\small
\setlength{\tabcolsep}{5pt}
\begin{tabular}{lccc}
\toprule
Benchmark & Budget & Score/Acc$\uparrow$ & Total Cost (\$)$\downarrow$ \\
\midrule
ScienceWorld & $0.5\times B$ & 45.2$\pm$3.8 & 3.6 \\
ScienceWorld & $1.0\times B$ & 53.8$\pm$3.2 & 5.7 \\
ScienceWorld & $2.0\times B$ & 55.6$\pm$3.1 & 8.1 \\
\midrule
HLE & $0.5\times B$ & 20.6$\pm$2.8 & 23.4 \\
HLE & $1.0\times B$ & 26.0$\pm$2.3 & 35.0 \\
HLE & $2.0\times B$ & 27.3$\pm$2.2 & 44.8 \\
\bottomrule
\end{tabular}
\caption{Budget sensitivity. Relaxing the budget improves performance, but with clear diminishing returns.}
\label{tab:budget_sweep}
\end{table}

\begin{table}[!htbp]
\centering
\scriptsize
\setlength{\tabcolsep}{3.5pt}
\resizebox{\linewidth}{!}{
\begin{tabular}{lcccccccc}
\toprule
& \multicolumn{2}{c}{ScienceWorld Test} & \multicolumn{2}{c}{ScienceWorld OOD} & \multicolumn{2}{c}{HLE Test} & \multicolumn{2}{c}{HLE OOD} \\
\cmidrule(lr){2-3}\cmidrule(lr){4-5}\cmidrule(lr){6-7}\cmidrule(lr){8-9}
Pool & Score$\uparrow$ & Cost$\downarrow$ & Score$\uparrow$ & Cost$\downarrow$ & Acc$\uparrow$ & Cost$\downarrow$ & Acc$\uparrow$ & Cost$\downarrow$ \\
\midrule
2-model & 50.6$\pm$3.6 & 8.9 & 7.2$\pm$4.3 & 28.5 & 25.4$\pm$2.5 & 45.2 & 36.4$\pm$3.1 & 43.5 \\
6-model & 53.8$\pm$3.2 & 5.7 & 9.9$\pm$3.9 & 16.3 & 26.0$\pm$2.3 & 35.0 & 38.6$\pm$3.0 & 31.2 \\
8-model & 54.1$\pm$3.3 & 5.3 & 10.3$\pm$4.1 & 16.8 & 25.8$\pm$2.4 & 34.2 & 39.1$\pm$2.9 & 29.5 \\
\bottomrule
\end{tabular}}
\caption{Candidate-pool sensitivity under 2-, 6-, and 8-model settings.}
\label{tab:pool_size}
\end{table}

\begin{table}[!htbp]
\centering
\small
\setlength{\tabcolsep}{6pt}
\begin{tabular}{lcc}
\toprule
History Budget & ScienceWorld$\uparrow$ & HLE$\uparrow$ \\
\midrule
\texttt{max\_tokens}=2048  & 49.3$\pm$3.7 & 23.8$\pm$2.6 \\
\texttt{max\_tokens}=4096  & 52.1$\pm$3.4 & 25.3$\pm$2.4 \\
\texttt{max\_tokens}=8192  & 53.8$\pm$3.2 & 26.0$\pm$2.3 \\
\texttt{max\_tokens}=16384 & 53.5$\pm$3.3 & 26.2$\pm$2.3 \\
\bottomrule
\end{tabular}
\caption{Sensitivity to the history token budget.}
\label{tab:history_budget}
\end{table}

\section{OpenRouter Baseline Details}
\label{app:openrouter}

\paragraph{Motivation.}
OpenRouter \citep{openrouter} provides an automatic routing API commonly used in commercial deployments.
We include OpenRouter as a representative commercial router baseline to contextualize \method against an off-the-shelf production routing system.

\paragraph{Model pool.}
Unlike our setting, which restricts routing to a fixed 6-model candidate pool (Table~\ref{tab:model_pool}), OpenRouter's automatic routing can choose from a much broader set of models.
In our evaluation, the OpenRouter baseline routes over the following pool (as reported by the API at routing time):

\begin{table*}[t]
\centering
\footnotesize
\begin{tabular}{ll}
\toprule
\textbf{Provider/Model} & \textbf{Provider/Model} \\
\midrule
\texttt{openai/gpt-5.1} & \texttt{openai/gpt-5} \\
\texttt{openai/gpt-5-mini} & \texttt{openai/gpt-5-nano} \\
\texttt{openai/gpt-4.1} & \texttt{openai/gpt-4.1-mini} \\
\texttt{openai/gpt-4.1-nano} & \texttt{openai/gpt-4o} \\
\texttt{openai/gpt-4o-2024-05-13} & \texttt{openai/gpt-4o-2024-08-06} \\
\texttt{openai/gpt-4o-2024-11-20} & \texttt{openai/gpt-4o-mini} \\
\texttt{openai/gpt-4o-mini-2024-07-18} & \texttt{openai/gpt-4-turbo} \\
\texttt{openai/gpt-4-turbo-preview} & \texttt{openai/gpt-4-1106-preview} \\
\texttt{openai/gpt-4} & \texttt{openai/gpt-3.5-turbo} \\
\texttt{openai/gpt-oss-120b} & \texttt{anthropic/claude-opus-4.5} \\
\texttt{anthropic/claude-opus-4.1} & \texttt{anthropic/claude-opus-4} \\
\texttt{anthropic/claude-sonnet-4.5} & \texttt{anthropic/claude-sonnet-4} \\
\texttt{anthropic/claude-3.7-sonnet} & \texttt{anthropic/claude-haiku-4.5} \\
\texttt{anthropic/claude-3.5-haiku} & \texttt{anthropic/claude-3-haiku} \\
\texttt{google/gemini-3-pro-preview} & \texttt{google/gemini-2.5-pro} \\
\texttt{google/gemini-2.0-flash-001} & \texttt{google/gemini-2.5-flash} \\
\texttt{mistralai/mistral-large} & \texttt{mistralai/mistral-large-2407} \\
\texttt{mistralai/mistral-large-2411} & \texttt{mistralai/mistral-medium-3.1} \\
\texttt{mistralai/mistral-nemo} & \texttt{mistralai/mistral-7b-instruct} \\
\texttt{mistralai/mixtral-8x7b-instruct} & \texttt{mistralai/mixtral-8x22b-instruct} \\
\texttt{mistralai/codestral-2508} & \texttt{x-ai/grok-4} \\
\texttt{x-ai/grok-3} & \texttt{x-ai/grok-3-mini} \\
\texttt{deepseek/deepseek-r1} & \texttt{meta-llama/llama-3.3-70b-instruct} \\
\texttt{meta-llama/llama-3.1-405b-instruct} & \texttt{meta-llama/llama-3.1-70b-instruct} \\
\texttt{meta-llama/llama-3.1-8b-instruct} & \texttt{meta-llama/llama-3-70b-instruct} \\
\texttt{meta-llama/llama-3-8b-instruct} & \texttt{qwen/qwen3-235b-a22b} \\
\texttt{qwen/qwen3-32b} & \texttt{qwen/qwen3-14b} \\
\texttt{cohere/command-r-plus-08-2024} & \texttt{cohere/command-r-08-2024} \\
\texttt{moonshotai/kimi-k2-thinking} & \texttt{perplexity/sonar} \\
\bottomrule
\end{tabular}
\caption{OpenRouter automatic routing model pool used by our OpenRouter baseline (as reported by the API at routing time).}
\label{tab:openrouter_pool}
\end{table*}

\paragraph{Implications for comparison.}
Because OpenRouter can select from a superset of models (including multiple frontier and provider-specific options not present in our pool), it represents a stronger routing setting than ours.
We still include it as a practical reference point, but comparisons should be interpreted with this mismatch in candidate pools in mind.

\begin{figure}[!htbp]
\centering
\includegraphics[width=\columnwidth]{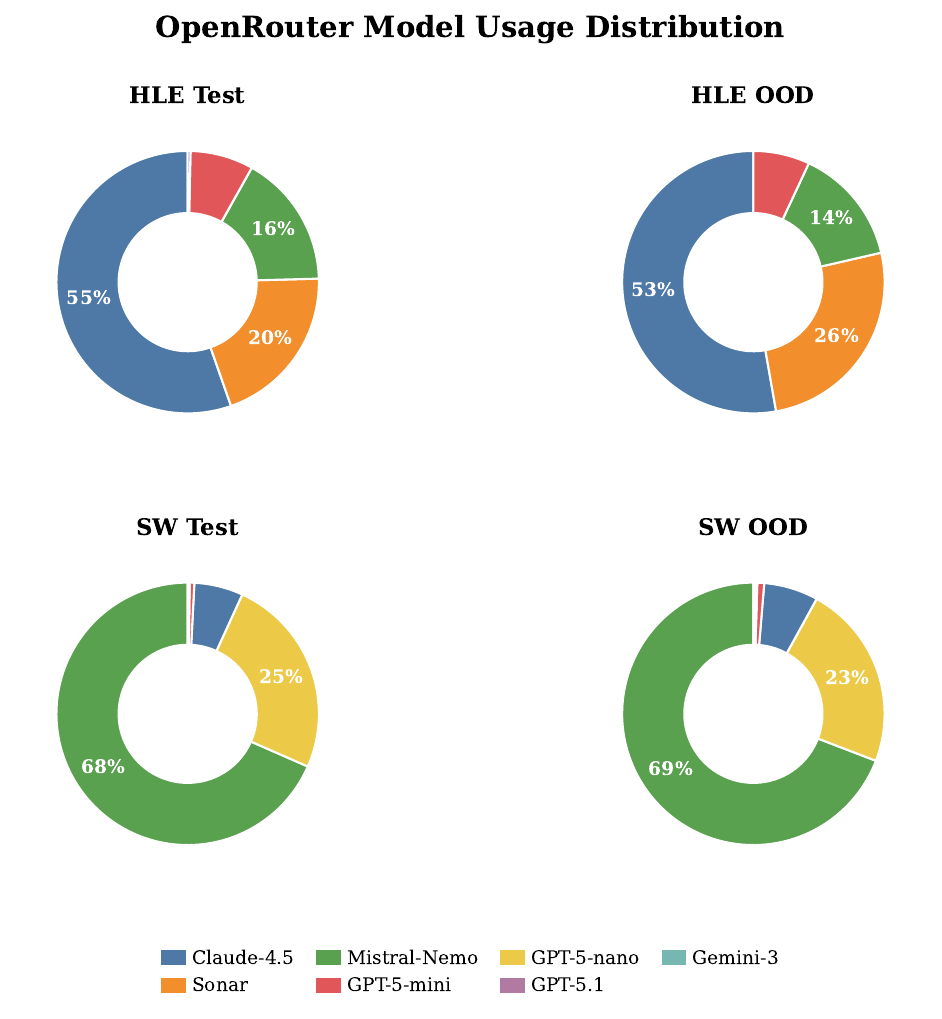}
\caption{Model usage distribution of OpenRouter on ScienceWorld and HLE. On ScienceWorld, OpenRouter predominantly selects lightweight models (Mistral-Nemo: 68\%, GPT-5-nano: 24\%), while on HLE it primarily uses stronger models (Claude-4.5: 54\%), reflecting different difficulty assessments.}
\label{fig:openrouter_model_usage}
\end{figure}

\paragraph{Why does OpenRouter perform poorly on ScienceWorld?}
Figure~\ref{fig:openrouter_model_usage} shows the model usage distribution of OpenRouter on both benchmarks.
On ScienceWorld, OpenRouter predominantly selects lightweight models: Mistral-Nemo accounts for 68\% of all model calls, followed by GPT-5-nano (24\%) and Claude-4.5 (7\%).
These models, while cost-efficient, lack the reasoning capability required for ScienceWorld's procedural scientific tasks.
In contrast, on HLE, OpenRouter allocates 54\% of calls to Claude-4.5---a much stronger frontier model---along with Sonar (23\%) and Mistral-Nemo (15\%), reflecting a more appropriate difficulty assessment for that benchmark.

This discrepancy reveals a key limitation of general-purpose commercial routers: without task-specific training, they may underestimate the difficulty of unfamiliar domains and over-rely on cheaper models.
ScienceWorld's text-based interface and seemingly simple commands may mislead the router into treating it as an easy task, when in fact it requires multi-step planning and precise action sequencing that weaker models struggle to execute.
The resulting negative scores ($-26.4$ on Test, $-26.9$ on OOD) indicate that the selected models frequently fail to make meaningful progress toward task completion, as judged by the environment's scoring function.

\section{Prompts and Tool Schemas}
\label{app:prompts}

We report the prompts used in our evaluation setup. For brevity, we omit the format-error correction prompts.

\subsection{ScienceWorld Agent Prompt}
\label{app:prompts:sw}
\begin{figure*}[!htbp]
\begin{paperbox}{ScienceWorld System Prompt}
You are a helpful assistant interacting with a text-based science simulation environment.
Your goal is to complete science experiments by issuing text commands.

\vspace{0.3cm}
\textbf{How to Interact} \\
You issue commands by writing them in a \texttt{```text} code block. The environment will respond with observations.

\vspace{0.2cm}
\texttt{<format\_example>} \\
\texttt{THOUGHT: I should explore my surroundings first.} \\
\texttt{```text} \\
\texttt{look around} \\
\texttt{```} \\
\texttt{</format\_example>}

\vspace{0.3cm}
\textbf{Available Command Types} \\
Common commands include:
\begin{itemize}
\item Movement: \texttt{go to [location]}, \texttt{open door}, \texttt{go through door}
\item Interaction: \texttt{pick up [object]}, \texttt{put [object] in [container]}, \texttt{activate [object]}
\item Observation: \texttt{look around}, \texttt{look at [object]}, \texttt{inventory}
\item Task-specific: \texttt{focus on [object]}, \texttt{use [tool] on [object]}
\end{itemize}

\vspace{0.2cm}
\textbf{Query Commands (Free Actions)} \\
You can query available actions without consuming a game turn:
\begin{itemize}
\item \texttt{?navigation}: show movement actions (go, walk, move)
\item \texttt{?object}: show object manipulation (pick up, put, pour)
\item \texttt{?observation}: show observation actions (look, examine, inventory)
\item \texttt{?device}: show device control (activate, turn on/off, use)
\item \texttt{?door}: show door/container actions (open, close)
\item \texttt{?electrical}: show electrical actions (connect, disconnect)
\item \texttt{?interaction}: show interaction actions (mix, eat, focus)
\item \texttt{?all}: show all valid actions
\item \texttt{?categories}: show query help
\end{itemize}
Use queries to explore available actions before deciding your next move.

\vspace{0.2cm}
\textbf{Important Notes}
\begin{itemize}
\item Issue ONE command per response
\item Include a THOUGHT section explaining your reasoning
\item The environment is turn-based---wait for observations before issuing the next command
\item Some tasks require multiple steps to complete
\end{itemize}
\end{paperbox}
\end{figure*}

\begin{figure*}[!htbp]
\begin{paperbox}{ScienceWorld Instance Prompt}
\texttt{<task\_description>} \\
\texttt{\{\{task\_description\}\}} \\
\texttt{</task\_description>}

\vspace{0.2cm}
\texttt{<initial\_observation>} \\
\texttt{\{\{initial\_observation\}\}} \\
\texttt{</initial\_observation>}

\vspace{0.2cm}
\texttt{<instructions>} \\
Complete the science experiment described above. You are interacting with a simulated environment.
Issue commands one at a time and observe the results.

\vspace{0.2cm}
\textbf{Tip}: Use query commands like \texttt{?navigation} or \texttt{?object} to explore available actions without consuming a turn.

\vspace{0.2cm}
\textbf{Strategy hints:}
\begin{itemize}
\item First explore: use \texttt{open door to [room]} then \texttt{go to [room]} to navigate
\item Look around each room to find objects you need
\item Pick up objects with \texttt{pick up [object]}
\item For heating: find a stove, turn it on with \texttt{activate [stove]}, place container on it
\item For cooling: use a freezer or refrigerator
\item Use \texttt{focus on [object]} to examine substances
\end{itemize}

\vspace{0.2cm}
To complete the task, perform the necessary actions described in the task description.
When you believe the task is complete, issue the command:

\vspace{0.2cm}
\texttt{```text} \\
\texttt{task completed} \\
\texttt{```}

\vspace{0.2cm}
\textbf{Remember:}
\begin{itemize}
\item Think step by step about what actions are needed
\item Explore your environment to find needed objects
\item Some actions may require prerequisites (e.g., picking up an object before using it)
\end{itemize}
\texttt{</instructions>}
\end{paperbox}
\end{figure*}

\subsection{HLE Agent Prompt}
\label{app:prompts:hle}
The HLE system prompt injects tool schemas via a template variable. We list each tool schema explicitly in Appendix~\ref{app:tool_schemas}.
\begin{figure*}[!htbp]
\begin{paperbox}{HLE System Prompt}
\small
You are an expert problem solver tackling challenging questions from Humanity's Last Exam.
These questions require deep reasoning, careful research, and precise answers.

\vspace{0.15cm}
\textbf{Available Tools} \\
\texttt{<tools>} \\
\texttt{\{\% for tool in tool\_schemas \%\}} \\
\texttt{\{\{ tool.to\_xml() \}\}} \\
\texttt{\{\% endfor \%\}} \\
\texttt{</tools>}

\vspace{0.15cm}
\textbf{Tool Call Formats} \\
You can use either XML or JSON format inside \texttt{<tool\_call>} tags:

\vspace{0.1cm}
\textbf{Format A: XML} \\
\texttt{<tool\_call>} \\
\texttt{<tool\_name>} \\
\texttt{\ \ <param1>value1</param1>} \\
\texttt{\ \ <param2>value2</param2>} \\
\texttt{</tool\_name>} \\
\texttt{</tool\_call>}

\vspace{0.1cm}
\textbf{Format B: JSON} \\
\texttt{<tool\_call>\{"name": "tool\_name", "arguments": \{"param1": "value1"\}\}</tool\_call>}

\vspace{0.15cm}
\textbf{Examples}

\vspace{0.05cm}
\texttt{<tool\_call>} \\
\texttt{<search>} \\
\texttt{\ \ <query>maximum likelihood estimation</query>} \\
\texttt{</search>} \\
\texttt{</tool\_call>}

\vspace{0.05cm}
\texttt{<tool\_call>\{"name": "python", "arguments": \{"code": "print(2 + 2)"\}\}</tool\_call>}

\vspace{0.15cm}
\textbf{Strategy for HLE Questions}
\begin{enumerate}[leftmargin=*, label=\arabic*., noitemsep, topsep=0pt]
\item \textbf{Understand the question}: read carefully and identify what's being asked
\item \textbf{Break down the problem}: decompose into sub-problems if needed
\item \textbf{Research if needed}: use \texttt{search} for factual information you're unsure about
\item \textbf{Verify sources}: use \texttt{browse} to read primary sources when accuracy matters
\item \textbf{Compute when needed}: use \texttt{python} for calculations and data analysis
\item \textbf{Synthesize}: combine information from multiple sources
\item \textbf{Verify your answer}: double-check before submitting
\item \textbf{Submit}: use \texttt{answer} with your final answer
\end{enumerate}

\vspace{0.1cm}
\textbf{Important Notes}
\begin{itemize}[noitemsep, topsep=0pt]
\item These are challenging questions---take your time
\item Be precise---exact answers are often required
\item Show your reasoning before using tools
\item If uncertain, express confidence level in your answer
\end{itemize}
\end{paperbox}
\end{figure*}

\begin{figure*}[!htbp]
\begin{paperbox}{HLE Instance Prompt}
\textbf{HLE Question} \\
\texttt{\{\% if subject \%\}} \\
\textbf{Subject}: \texttt{\{\{ subject \}\}} \\
\texttt{\{\% endif \%\}}

\vspace{0.2cm}
\texttt{\{\{ question | default(task) \}\}}

\vspace{0.2cm}
\hrule
\vspace{0.2cm}

Solve this step by step. Use tools to research and verify.
Submit your final answer using the \texttt{answer} tool:
\texttt{<tool\_call>\{"name": "answer", "arguments": \{"answer": "your final answer"\}\}</tool\_call>}
\end{paperbox}
\end{figure*}

\subsection{Tool Schemas}
\label{app:tool_schemas}
We list each tool's JSON schema (name, description, and parameter schema) used by the HLE agent.
\begin{figure*}[!htbp]
\begin{tcolorbox}[
    enhanced,
    title=\textbf{Tool Schema: search},
    colback=white,
    colframe=black!90,
    coltitle=white,
    fonttitle=\bfseries\small,
    boxrule=0.8pt,
    arc=2mm,
    attach boxed title to top left={xshift=4mm, yshift=-3mm},
    boxed title style={colback=black!90, sharp corners=south}
]
\begin{lstlisting}[language=json]
{
  "name": "search",
  "description": "Search the web for information using Google",
  "parameters": {
    "type": "object",
    "properties": {
      "query": {
        "oneOf": [
          { "type": "string" },
          { "type": "array", "items": { "type": "string" } }
        ],
        "description": "Search query or list of queries for batch search"
      },
      "num_results": {
        "type": "integer",
        "description": "Number of results to return (default: 10)",
        "minimum": 1,
        "maximum": 100
      }
    }
  },
  "required": [ "query" ]
}
\end{lstlisting}
\end{tcolorbox}
\end{figure*}

\begin{figure*}[!htbp]
\begin{tcolorbox}[
    enhanced,
    title=\textbf{Tool Schema: browse},
    colback=white,
    colframe=black!90,
    coltitle=white,
    fonttitle=\bfseries\small,
    boxrule=0.8pt,
    arc=2mm,
    attach boxed title to top left={xshift=4mm, yshift=-3mm},
    boxed title style={colback=black!90, sharp corners=south}
]
\begin{lstlisting}[language=json]
{
  "name": "browse",
  "description": "Visit webpage(s) and extract information based on a goal",
  "parameters": {
    "type": "object",
    "properties": {
      "url": {
        "type": "string",
        "description": "URL to visit. Can be a single URL or array of URLs."
      },
      "goal": {
        "type": "string",
        "description": "The goal of the visit - what information to extract from the webpage(s)."
      },
      "extract_mode": {
        "type": "string",
        "enum": [ "text", "markdown", "html" ],
        "description": "Content extraction mode (default: text). Only used when LLM summary is disabled."
      }
    }
  },
  "required": [ "url", "goal" ]
}
\end{lstlisting}
\end{tcolorbox}
\end{figure*}

\begin{figure*}[!htbp]
\begin{tcolorbox}[
    enhanced,
    title=\textbf{Tool Schema: python},
    colback=white,
    colframe=black!90,
    coltitle=white,
    fonttitle=\bfseries\small,
    boxrule=0.8pt,
    arc=2mm,
    attach boxed title to top left={xshift=4mm, yshift=-3mm},
    boxed title style={colback=black!90, sharp corners=south}
]
\begin{lstlisting}[language=json]
{
  "name": "python",
  "description": "Execute Python code and return the output",
  "parameters": {
    "type": "object",
    "properties": {
      "code": {
        "type": "string",
        "description": "Python code to execute"
      }
    }
  },
  "required": [ "code" ]
}
\end{lstlisting}
\end{tcolorbox}
\end{figure*}

\begin{figure*}[!htbp]
\begin{tcolorbox}[
    enhanced,
    title=\textbf{Tool Schema: answer},
    colback=white,
    colframe=black!90,
    coltitle=white,
    fonttitle=\bfseries\small,
    boxrule=0.8pt,
    arc=2mm,
    attach boxed title to top left={xshift=4mm, yshift=-3mm},
    boxed title style={colback=black!90, sharp corners=south}
]
\begin{lstlisting}[language=json]
{
  "name": "answer",
  "description": "Submit your final answer to complete the task",
  "parameters": {
    "type": "object",
    "properties": {
      "answer": {
        "type": "string",
        "description": "The final answer to submit"
      },
      "confidence": {
        "type": "number",
        "description": "Confidence level from 0 to 1 (optional)",
        "minimum": 0,
        "maximum": 1
      },
      "reasoning": {
        "type": "string",
        "description": "Brief explanation of how you arrived at the answer (optional)"
      }
    }
  },
  "required": [ "answer" ]
}
\end{lstlisting}
\end{tcolorbox}
\end{figure*}

\subsection{Routing Model Prompt for LLM Router / Router-R1}
\label{app:prompts:routing}
We report the turn-level routing prompt used by the LLM Router baseline and by Router-R1. In both baselines, the routing model is queried at each turn to produce a \texttt{<select>} decision. They use the same prompt template; Router-R1 uses a trained \texttt{Qwen2.5-7B-Instruct} routing model, while LLM Router directly uses \texttt{DeepSeek-V3.2} (no training).
\begin{figure*}[!htbp]
\begin{paperbox}{LLM Router / Router-R1 System Prompt (Policy LLM)}
You are a model routing assistant. Your job is to select the best model for the given task.

\vspace{0.3cm}
\textbf{Available models:} \\
\texttt{\{model\_list\}}

\vspace{0.3cm}
\textbf{Instructions}
\begin{enumerate}[leftmargin=*, label=\arabic*.]
\item Analyze the task in \texttt{<think>...</think>} tags
\item Select the best model using \texttt{<select>model\_name</select>}
\end{enumerate}

\vspace{0.2cm}
\textbf{Example} \\
\texttt{<think>This is a simple math question. A cheaper model would suffice.</think>} \\
\texttt{<select>deepseek/deepseek-v3.2</select>}

\vspace{0.3cm}
\textbf{Rules}
\begin{itemize}
\item You MUST output exactly one \texttt{<select>} tag
\item The model name must match exactly from the available list
\item Consider: task complexity, model strengths, cost-effectiveness
\end{itemize}
\end{paperbox}
\end{figure*}

\begin{figure*}[!htbp]
\begin{paperbox}{LLM Router / Router-R1 Model Descriptors}
\texttt{openai/gpt-5}: Cost 1.25/10. 400K context. Top performer with SW 48\% SR and HLE 25\% SR. Best for complex multi-step reasoning and hard science problems. \\
\texttt{openai/gpt-oss-120b}: Cost 0.09/0.36. 131K context. Lowest cost with SW 33\% SR and HLE 10\% SR. Excellent cost-efficiency for moderate difficulty tasks. \\
\texttt{moonshotai/kimi-k2-0905}: Cost 0.39/1.9. 131K context. SW 31\% SR and HLE 11\% SR. Balanced option for general-purpose tasks with good instruction following. \\
\texttt{deepseek/deepseek-v3.2}: Cost 0.27/0.42. 164K context. SW 16\% SR and HLE 16\% SR. Strong on math and coding. Good for structured reasoning tasks. \\
\texttt{google/gemini-2.5-flash-lite}: Cost 0.1/0.4. 1M context. SW 23\% SR and HLE 6\% SR. Largest context window. Best for long document analysis. \\
\texttt{minimax/minimax-m2}: Cost 0.2/1. 196K context. SW 22\% SR and HLE 8\% SR. Fast inference. Suitable for simple QA and straightforward tasks.
\end{paperbox}
\end{figure*}

\subsection{Browse Extractor Prompt}
\label{app:prompts:browse}
The HLE \texttt{browse} tool can optionally use an LLM to extract and summarize relevant evidence from retrieved webpages. We report the extractor prompt used for this LLM-based summarization.
\begin{figure*}[!htbp]
\begin{paperbox}{Browse Extractor Prompt (LLM Summary)}
Please process the following webpage content and user goal to extract relevant information:

\vspace{0.3cm}
\textbf{\large Webpage Content} \\
\{webpage\_content\}

\vspace{0.3cm}
\textbf{\large User Goal} \\
\{goal\}

\vspace{0.3cm}
\textbf{\large Task Guidelines}
\begin{enumerate}[leftmargin=*, label=\arabic*.]
\item \textbf{Content Scanning for Rationale}: Locate the \textbf{specific sections/data} directly related to the user's goal within the webpage content
\item \textbf{Key Extraction for Evidence}: Identify and extract the \textbf{most relevant information} from the content; do not miss important information. Output the \textbf{full original context} as much as possible (it can exceed three paragraphs).
\item \textbf{Summary Output for Summary}: Organize into a concise paragraph with logical flow, prioritizing clarity and judging the contribution of the information to the goal.
\end{enumerate}

\vspace{0.2cm}
\textbf{Output Format}: JSON format containing \texttt{"rational"}, \texttt{"evidence"}, and \texttt{"summary"} fields.
\end{paperbox}
\end{figure*}

\subsection{HLE Judge Prompt}
\label{app:prompts:judge}
HLE scoring uses an LLM-as-a-judge component aligned with the official HLE evaluation. We report the judge prompt used to compare a model response against the provided reference answer.
\begin{figure*}[!htbp]
\begin{paperbox}{HLE Judge Prompt}
Judge whether the following \texttt{[response]} to \texttt{[question]} is correct or not based on the precise and unambiguous \texttt{[correct\_answer]} below.

\vspace{0.3cm}
\texttt{[question]: \{question\}} \\
\texttt{[response]: \{response\}}

\vspace{0.3cm}
Your judgement must be in the format and criteria specified below:

\vspace{0.2cm}
\textbf{extracted\_final\_answer}: The final exact answer extracted from the \texttt{[response]}. Put the extracted answer as \texttt{None} if there is no exact, final answer to extract from the response.

\vspace{0.2cm}
\texttt{[correct\_answer]: \{correct\_answer\}}

\vspace{0.2cm}
\textbf{reasoning}: Explain why the extracted\_final\_answer is correct or incorrect based on \texttt{[correct\_answer]}, focusing only on whether there are meaningful differences. Do not comment on background, do not attempt to solve the problem, do not argue for any answer different than \texttt{[correct\_answer]}; focus only on whether the answers match.

\vspace{0.2cm}
\textbf{correct}: Answer \texttt{yes} if extracted\_final\_answer matches the \texttt{[correct\_answer]} given above, or is within a small margin of error for numerical problems. Answer \texttt{no} otherwise.

\vspace{0.2cm}
\textbf{confidence}: The extracted confidence score between 0\% and 100\% from \texttt{[response]}. Put 100 if there is no confidence score available.

\vspace{0.2cm}
Respond with a JSON object containing these four fields: extracted\_final\_answer, reasoning, correct, confidence.
\end{paperbox}
\end{figure*}

\end{document}